\documentclass[a4paper,twocolumn,11pt,unpublished]{quantumarticle}
\pdfoutput=1
\usepackage[utf8]{inputenc}
\usepackage[english]{babel}
\usepackage[T1]{fontenc}
\usepackage[pagebackref]{hyperref}

\usepackage{tikz}
\usetikzlibrary{positioning}

\usepackage[numbers]{natbib}

\usepackage{makecell}
\usepackage[table]{xcolor}
\usepackage{hhline}

\usepackage{xurl}

\usepackage{booktabs}
\usepackage{multirow}

\usepackage{fancyvrb}
\usepackage{threeparttable}

\fvset{frame=single,framesep=1mm,fontfamily=courier,fontsize=\scriptsize,numbers=left,framerule=.3mm,numbersep=1mm,commandchars=\\\{\}}

\usetikzlibrary{shapes.geometric, fit}

\begin{document}

\title{Revisiting Quantum Code Generation: Where Should Domain Knowledge Live?}

\author{Oscar Novo}
\affiliation{Quantum Computing Research, QCentroid, 48001, Bilbao, Spain}
\orcid{0000-0002-5123-2608}

\author{Oscar Bastidas-Jossa}
\affiliation{Quantum Computing Research, QCentroid, 48001, Bilbao, Spain}
\orcid{0000-0003-4806-5754}

\author{Alberto Calvo}
\affiliation{Quantum Computing Research, QCentroid, 48001, Bilbao, Spain}
\orcid{0009-0007-9579-5342}

\author{Antonio Peris}
\affiliation{Quantum Computing Research, QCentroid, 48001, Bilbao, Spain}
\orcid{0009-0002-1333-9233}

\author{Carlos Kuchkovsky}
\affiliation{Quantum Computing Research, QCentroid, 48001, Bilbao, Spain}
\orcid{0000-0002-2834-3967}

\maketitle

\begin{abstract}

Recent advances in large language models (LLMs) have enabled the automation of an increasing number of programming tasks, including code generation for scientific and engineering domains. In rapidly evolving software ecosystems such as quantum software development, where frameworks expose complex abstractions, a central question is how best to incorporate domain knowledge into LLM-based assistants while preserving maintainability as libraries evolve.

In this work, we study \emph{specialization strategies} for Qiskit code generation using the Qiskit-HumanEval benchmark. We compare a \emph{parameter-specialized} fine-tuned baseline introduced in prior work against a range of recent general-purpose LLMs enhanced with retrieval-augmented generation (RAG) and agent-based inference with execution feedback.

Our results show that modern general-purpose LLMs consistently outperform the parameter-specialized baseline. While the fine-tuned model achieves approximately 47\% pass@1 on Qiskit-HumanEval, recent general-purpose models reach 60--65\% under zero-shot and retrieval-augmented settings, and up to 85\% for the strongest evaluated model when combined with iterative execution-feedback agents---representing an improvement of more than 20\% over zero-shot general-purpose performance and more than 35\% over the parameter-specialized baseline.

Agentic execution feedback yields the most consistent improvements, albeit at increased runtime cost, while RAG provides modest and model-dependent gains. These findings indicate that performance gains can be achieved without domain-specific fine-tuning, instead relying on inference-time augmentation, thereby enabling a more flexible and maintainable approach to LLM-assisted quantum software development.
\end{abstract}

\section{Introduction}

Large language models (LLMs) \cite{vaswani2023attentionneed} have recently emerged as powerful tools for automating a wide range of programming tasks, including code completion, program synthesis, and software maintenance. Their impact has been particularly notable in scientific and engineering domains, where LLMs are increasingly used to assist with complex workflows involving specialized libraries, numerical methods, and domain-specific abstractions. As these models continue to improve in scale, reasoning capability, and context length \cite{chen2021codex}, an important open question is how best to incorporate domain knowledge into LLM-based systems for specialized scientific software ecosystems.

In quantum computing, the development of reliable and efficient software remains a key challenge, given the complexity of quantum programming abstractions and the rapid evolution of quantum software development kits (SDKs), such as Qiskit \cite{javadiabhari2024quantumcomputingqiskit}, Cirq \cite{Cirq_Developers_2025}, PennyLane \cite{bergholm2022pennylaneautomaticdifferentiationhybrid}, and Braket SDK \cite{amazonbraket}. These SDKs expose diverse programming abstractions, execution backends, and rapidly evolving APIs, making correctness and maintainability central concerns for automated code generation.

Broadly, domain adaptation for LLM-based code generation can be approached through two complementary paradigms. The first is \emph{training-time parameter-level specialization}, in which a base model is fine-tuned on curated domain corpora to encode framework-specific knowledge in its weights. The second, which we refer to as \emph{inference-time system-level specialization}, adapts a general-purpose model at inference time by embedding it within a structured pipeline that incorporates retrieval mechanisms and agent-based execution-feedback loops. This approach enables context-aware code generation by incorporating external retrieval and execution feedback, supporting iterative self-correction without modifying the model's parameters.

Prior work demonstrated that training-time specialization can yield substantial improvements for Qiskit code generation by fine-tuning a Granite-based model on curated Qiskit corpora and evaluating on the Qiskit-HumanEval benchmark \cite{dupuis2024qiskitcodeassistanttraining,qiskit_humaneval_2024}. However, the landscape of general-purpose LLM capabilities has evolved rapidly, with notable improvements in long-context reasoning, code synthesis, and tool use. At the same time, training-time specialization entails non-trivial costs, including data curation, training infrastructure, maintenance overhead, and limited adaptability to evolving software libraries. These considerations motivate a reassessment of the trade-offs between parameter-level specialization and inference-time system-level approaches.

In this work, we revisit Qiskit code generation and systematically re-evaluate performance on Qiskit-HumanEval \cite{qiskit_humaneval_2024} using a range of recent general-purpose LLMs, alongside a parameter-specialized fine-tuned baseline. Beyond direct model comparisons, we study inference-time augmentation strategies, including retrieval-augmented generation (RAG) \cite{lewis2021retrievalaugmentedgenerationknowledgeintensivenlp} and agent-based workflows \cite{yao2023reactsynergizingreasoningacting} that incorporate execution feedback. These techniques enable models to dynamically access external knowledge and iteratively refine generated code without modifying model parameters, offering a flexible alternative to static specialization.

Our study aims to answer the following research questions: (i) Do modern general-purpose LLMs match or exceed the performance of a parameter-specialized baseline on Qiskit code generation benchmarks? (ii) To what extent can retrieval and agentic inference strategies improve performance without additional training? and (iii) What are the implications of these findings for the design, cost, and long-term maintainability of LLM-based systems for quantum software development?

The contributions of this paper are threefold. First, we provide an updated empirical evaluation of Qiskit code generation performance in the context of recent general-purpose LLMs using Qiskit-HumanEval. Second, we analyze the effectiveness of retrieval-augmented and agent-based inference strategies for Qiskit code generation. Third, we discuss broader implications with respect to reproducibility, benchmark leakage risk, cost-efficiency, and the sustainability of training-time specialization approaches.

\section{Related Work}

The most closely related prior work to this study is the Qiskit Code Assistant study introduced by Dupuis et al.~\cite{dupuis2024qiskitcodeassistanttraining}. In that work, the authors propose a domain-specific approach to quantum code generation by fine-tuning a Granite-based large language model on curated Qiskit-related corpora. To evaluate the effectiveness of this strategy, they introduce the Qiskit-HumanEval benchmark \cite{qiskit_humaneval_2024}, a quantum programming benchmark derived from HumanEval \cite{chen2021codex}, and report substantial improvements over general-purpose code models on Qiskit-centric tasks.

Table~\ref{tab:he_qhe} reproduces results reported by Dupuis et al.~\cite{dupuis2024qiskitcodeassistanttraining}, comparing general-purpose code models, instruction-tuned variants, and domain-specific fine-tuned models. In particular, the Qiskit-fine-tuned Granite-20B model shows a large performance gain on Qiskit-HumanEval relative to the base Granite model and several general-purpose LLMs, motivating domain-specific fine-tuning as an effective strategy for quantum code generation at the time of publication.

\begin{table}[t]
\centering
\small
\setlength{\tabcolsep}{10pt}
\renewcommand{\arraystretch}{1.08}
\caption{HumanEval (HE) and Qiskit-HumanEval (QHE) pass@1 results reported by Dupuis et al.~\cite{dupuis2024qiskitcodeassistanttraining}. Pass@1 denotes the fraction of tasks for which a single generated solution passes all unit tests. Model names are simplified for clarity.}
\label{tab:he_qhe}
\begin{tabular}{l|c|c}
\hline
\textbf{Model} & \textbf{HE} & \textbf{QHE} \\
\hline
\makecell[l]{CodeLLaMA-34B\\(Python-Specialized)}        & 52.43\% & 26.73\% \\
\arrayrulecolor{gray!35}\hhline{---}\arrayrulecolor{black}
\makecell[l]{DeepSeek-Coder-33B\\(Base)}                 & 49.39\% & 39.60\% \\
\arrayrulecolor{gray!35}\hhline{---}\arrayrulecolor{black}
\makecell[l]{DeepSeek-Coder-33B\\(Instruction-Tuned)}    & \textbf{68.9\%} & 35.64\% \\
\arrayrulecolor{gray!35}\hhline{---}\arrayrulecolor{black}
StarCoder2-15B                                           & 45.12\% & 37.62\% \\
\arrayrulecolor{gray!35}\hhline{---}\arrayrulecolor{black}
\makecell[l]{Granite-20B\\(Base Code Model)}             & 38.41\% & 20.79\% \\
\arrayrulecolor{gray!35}\hhline{---}\arrayrulecolor{black}
\makecell[l]{\textbf{Granite-20B}\\\textbf{(Qiskit-Fine-Tuned)}} & 36.58\% & \textbf{46.53\%} \\
\hline
\end{tabular}
\end{table}

While these findings provide strong evidence for the effectiveness of parameter-level specialization, the evaluation predates recent advances in general-purpose LLM capabilities and does not consider inference-time augmentation strategies such as retrieval-augmented generation (RAG) \cite{lewis2021retrievalaugmentedgenerationknowledgeintensivenlp} or agent-based execution feedback \cite{yao2023reactsynergizingreasoningacting}. As a result, it remains unclear whether similar gains can be achieved using modern general-purpose models combined with enhanced inference techniques, without resorting to domain-specific fine-tuning.

\subsection{Inference-Time Augmentation for Code Generation}

Beyond parameter-level specialization, recent work has explored inference-time augmentation techniques that enhance the capabilities of large language models without additional training. One prominent approach is retrieval-augmented generation (RAG), which couples an LLM with an external retrieval component to dynamically incorporate relevant information—such as documentation, API references, or source code—into the model’s input context during generation~\cite{lewis2021retrievalaugmentedgenerationknowledgeintensivenlp}. By grounding generation in retrieved context, RAG reduces reliance on memorized knowledge and has been shown to improve performance on knowledge-intensive and API-heavy coding tasks \cite{lewis2021retrievalaugmentedgenerationknowledgeintensivenlp}.

Complementary to retrieval-based methods, agent-based inference frameworks extend LLMs with the ability to iteratively reason, invoke tools, execute code, and incorporate feedback in closed-loop systems~\cite{yao2023reactsynergizingreasoningacting}. In the context of code generation, such agentic approaches enable models to refine outputs based on execution results, diagnose errors, and perform multi-step problem solving, often yielding higher success rates on complex programming benchmarks.

A key advantage of inference-time augmentation methods is that they operate without modifying model parameters. As a result, the same retrieval or agentic framework can be applied across different base models and model versions, improving flexibility and reducing the cost and maintenance overhead associated with retraining or fine-tuning. This model-agnostic nature is particularly appealing in settings where underlying LLMs evolve rapidly or where multiple models are evaluated or used in parallel.

These inference-time strategies differ fundamentally from instruction tuning, which improves instruction-following behavior through supervised fine-tuning on generic prompt--response data but does not adapt models to specific software frameworks or APIs. In contrast, retrieval-augmented and agent-based methods enhance model performance at inference time by providing external context and structured interaction with execution environments.

While retrieval-augmented and agentic techniques have demonstrated promising results across a range of programming and reasoning benchmarks, their effectiveness in quantum programming remains largely unexplored. This is particularly relevant given the strict semantic correctness requirements and rapidly evolving APIs characteristic of quantum software frameworks.

In this work, we revisit Qiskit code generation with modern general-purpose LLMs and evaluate whether inference-time system-level specialization can replace training-time fine-tuning for this benchmark.

\section{Experimental Setup}

We evaluated quantum code generation using the Qiskit-HumanEval benchmark, a domain-specific adaptation of the HumanEval framework~\cite{chen2021codex} for quantum programming, obtained from the official open-source repository maintained by the Qiskit community~\cite{qiskit_humaneval_2024}. The benchmark comprises programming tasks that require generating functionally correct Qiskit code from natural-language specifications, with correctness assessed through executable unit tests.

Each task follows the standard HumanEval structure and is represented as a JSON object containing a prompt (including the function signature and docstring), an entry-point function name, and a Python test harness used to validate the generated solution. During evaluation, the model-generated code is inserted into a predefined template and executed in a sandboxed Python environment. A task is considered successful if all assertions in the test harness pass without raising an exception; any runtime error, incorrect return type, or failed assertion results in a failed task. An illustrative example from the Qiskit-HumanEval benchmark is shown below.

\begin{center}
\begin{minipage}{\linewidth}
\begin{Verbatim}[label={Qiskit-HumanEval task Sample}]
task_id: qiskitHumanEval/0,
prompt: from qiskit import QuantumCircuit
         def create_quantum_circuit(n_qubits):
    \textbf{"Generate a QuantumCircuit with n_qubits",}
canonical_solution: 
      return QuantumCircuit(n_qubits),
test: def check(candidate):
       result = candidate(3)
       assert isinstance(result, QuantumCircuit)
       assert result.num_qubits == 3,
entry_point: create_quantum_circuit,
difficulty_scale: basic
\end{Verbatim}
\end{minipage}
\end{center}

To ensure comparability with prior work, we adopt the same benchmark version used in the original Qiskit Code Assistant study~\cite{dupuis2024qiskitcodeassistanttraining}. Unless otherwise stated, results are reported using the standard HumanEval \emph{pass@1} metric~\cite{chen2021codex}, in which a single solution is generated per task and the fraction of tasks whose solution passes all associated unit tests is reported. When k = 1, Pass@1 corresponds to task-level accuracy.

\paragraph{Models.}
As a training-time specialization baseline, we use the Qiskit-focused fine-tuned Granite model reported by Dupuis et al.~\cite{dupuis2024qiskitcodeassistanttraining}. We refer to this model as the \emph{parameter-specialized baseline} throughout the paper.

We compare this baseline against a diverse set of general-purpose frontier LLMs accessed via commercial APIs, including OpenAI models (general-purpose and reasoning-oriented variants) \cite{openai_models_2025}, Google Gemini models \cite{geminiteam2025geminifamilyhighlycapable}, and Anthropic Claude models \cite{anthropic_claude_models_2025}. These models are not fine-tuned for quantum programming and are evaluated in an out-of-the-box configuration. Model versions and access modalities are explicitly recorded to ensure temporal traceability.

We include a single parameter-specialized baseline because it is currently the only publicly reported fine-tuned Qiskit model evaluated on Qiskit-HumanEval under a comparable protocol \cite{dupuis2024qiskitcodeassistanttraining}.

\paragraph{Evaluation methodology.}

All models are evaluated under identical task prompts and test conditions. Performance is reported using the standard HumanEval \emph{pass@1} metric, obtained from a single evaluation run, with one generated solution per task. To assess robustness under stochastic inference, we repeat the full evaluation five times and verify that the resulting \emph{pass@1} scores are consistent. Decoding parameters are kept fixed across models and runs unless explicitly stated.

Retrieval-based and agentic approaches introduce additional inference-time overhead compared to fine-tuned models. While inference runtime measurements are not reported in the original evaluation study of the parameter-specialized baseline \cite{dupuis2024qiskitcodeassistanttraining}, we include inference runtime measurements for general-purpose LLMs to contextualize the practical trade-offs associated with retrieval-augmented and agentic configurations across models.

\subsection{Inference Configurations}
\label{sec:methods}

To isolate the impact of parameter-level specialization versus inference-time augmentation, each general-purpose model is evaluated under three inference configurations: (i) zero-shot generation, where the model receives only the task prompt; (ii) retrieval-augmented generation (RAG) \cite{lewis2021retrievalaugmentedgenerationknowledgeintensivenlp}, where relevant Qiskit documentation is retrieved and added to the prompt as additional context; and (iii) agent-based inference \cite{yao2023reactsynergizingreasoningacting}, which builds upon retrieval augmentation by incorporating execution errors into an iterative feedback loop to guide subsequent repair attempts. For agent-based inference, we evaluate multiple configurations with a bounded number of repair iterations, ranging from a single retry to a maximum of five iterations. The domain-specific fine-tuned Granite model is evaluated in its standard inference configuration, without retrieval or agentic augmentation, consistent with the original evaluation protocol reported in prior work \cite{dupuis2024qiskitcodeassistanttraining}.

Prior to the full benchmark evaluation, we performed exploratory experiments to assess the impact of retrieval augmentation. The final inference configuration was selected based on overall performance and runtime considerations. This configuration was then applied uniformly across all evaluated general-purpose models.

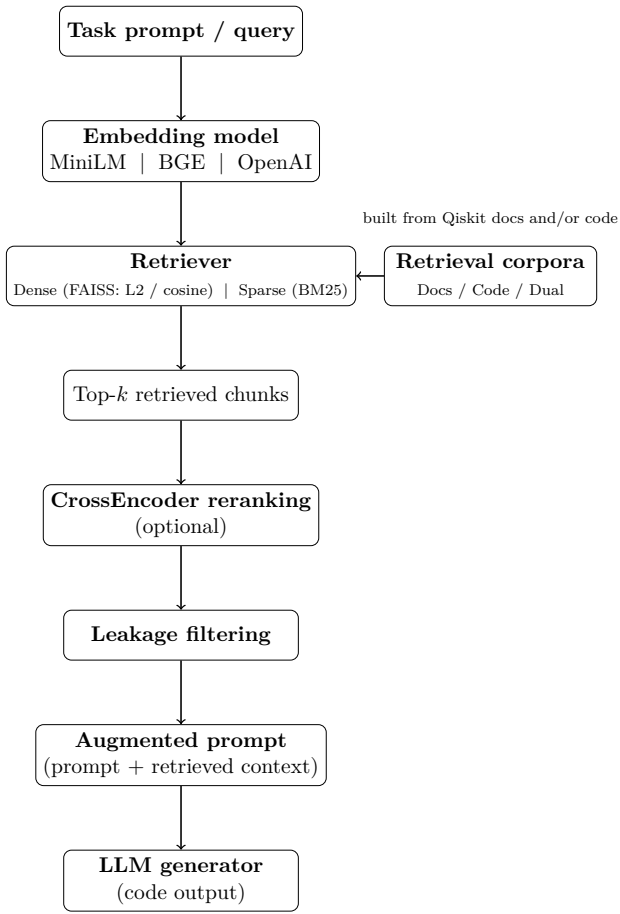
\begin{figure}[t]
\centering
\resizebox{1\columnwidth}{!}{%
\begin{tikzpicture}[
  node distance=1.15cm,
  box/.style={draw, rounded corners, align=center, minimum width=4.2cm, minimum height=0.9cm},
  sidebox/.style={draw, rounded corners, align=center, minimum width=3.8cm, minimum height=0.9cm},
  arrow/.style={->, thick},
  note/.style={font=\scriptsize, align=center}
]

\node[box] (query) {\textbf{Task prompt / query}};
\node[box, below=of query] (embed) {\textbf{Embedding model}\\
MiniLM \;|\; BGE \;|\; OpenAI};
\node[box, below=of embed] (retrieve) {\textbf{Retriever}\\
\scriptsize Dense (FAISS: L2 / cosine) \;|\; Sparse (BM25)};
\node[box, below=of retrieve] (topk) {Top-$k$ retrieved chunks};
\node[box, below=of topk] (rerank) {\textbf{CrossEncoder reranking} \\(optional)};
\node[box, below=of rerank] (filter) {\textbf{Leakage filtering}};
\node[box, below=of filter] (aug) {\textbf{Augmented prompt}\\(prompt + retrieved context)};
\node[box, below=of aug] (llm) {\textbf{LLM generator} \\ (code output)};

\node[sidebox, right=0.5cm of retrieve] (index) {\textbf{Retrieval corpora}\\\scriptsize Docs / Code / Dual};

\draw[arrow] (query) -- (embed);
\draw[arrow] (embed) -- (retrieve);
\draw[arrow] (retrieve) -- (topk);
\draw[arrow] (topk) -- (rerank);
\draw[arrow] (rerank) -- (filter);
\draw[arrow] (filter) -- (aug);
\draw[arrow] (aug) -- (llm);

\draw[arrow] (index) -- (retrieve);

\node[note, above=0.2cm of index] {\scriptsize built from Qiskit docs and/or code};

\end{tikzpicture}%
}
\caption{Simplified RAG pipeline. A query is embedded and used to retrieve relevant chunks from corpora built over Qiskit documentation and/or source code (using dense FAISS or sparse BM25 retrieval), optionally reranked, and incorporated into the prompt for code generation.}
\label{fig:rag_pipeline}
\end{figure}

\paragraph{Retrieval-augmented generation (RAG).}
For retrieval-augmented generation (RAG) experiments, we explored a range of retrieval and embedding configurations to assess their effectiveness in the quantum programming setting (see Figure~\ref{fig:rag_pipeline}).

As summarized in Table~\ref{tab:embedding_models}, we evaluated four \textbf{embedding models} with varying size and computational characteristics. Embedding models map text---including documentation, source code, and natural-language queries---into high-dimensional vector representations that capture semantic similarity and enable efficient similarity-based retrieval. The models were the following: \textit{all-MiniLM-L6-v2} (22M parameters, CPU-friendly), \textit{BAAI/bge-base-en-v1.5} (109M parameters), \textit{BAAI/bge-large-en-v1.5} (335M parameters)\cite{bge_v1_v15_documentation}, and \textit{text-embedding-3-large} (3--5B parameter range, cloud-based) \cite{openai_embeddings_2025}. The BAAI and OpenAI embeddings are GPU-recommended, while \textit{all-MiniLM-L6-v2} supports lightweight CPU-only retrieval.

\begin{table}[t]
\centering
\footnotesize
\begin{threeparttable}
\caption{Embedding models evaluated as retrieval backbones.}
\label{tab:embedding_models}

\begin{tabular}{@{}l c l@{}}
\toprule
\textbf{Model} & \textbf{Params} & \textbf{Compute / Deploy.} \\
\midrule
\makecell[l]{all-MiniLM-L6-v2} 
& 22M 
& \makecell[l]{CPU-friendly;\\CPU-only feasible} \\
\arrayrulecolor{gray!35}\hhline{---}\arrayrulecolor{black}
\makecell[l]{BAAI/bge-base-en-v1.5} 
& 109M 
& \makecell[l]{GPU recommended} \\
\arrayrulecolor{gray!35}\hhline{---}\arrayrulecolor{black}
\makecell[l]{BAAI/bge-large-en-v1.5} 
& 335M 
& \makecell[l]{GPU recommended} \\
\arrayrulecolor{gray!35}\hhline{---}\arrayrulecolor{black}
\makecell[l]{text-embedding-3-large} 
& \makecell[c]{$\sim$3-5B\\range} 
& \makecell[l]{Cloud-based;\\GPU recommended\\(provider-side)} \\
\bottomrule
\end{tabular}

\vspace{3pt}
\footnotesize
\textit{Note:} BAAI and OpenAI embeddings are GPU-recommended, while all-MiniLM-L6-v2 supports lightweight CPU-only retrieval.
\end{threeparttable}
\end{table}

For retrieval, we investigated both \textbf{dense} and \textbf{sparse} approaches, as well as different similarity and ranking strategies. Dense retrieval is implemented using \textbf{FAISS} \cite{douze2024faiss}, a high-performance vector similarity search library that indexes embedding vectors and retrieves documents based on nearest-neighbor search in vector space. We primarily used the L2 (Euclidean distance) metric for dense retrieval. Sparse retrieval was implemented using \textbf{BM25} \cite{robertson2009probabilistic}, a frequency--inverse document frequency (TF--IDF) based method that emphasizes exact keyword overlap; while effective for precise term matching, it is less capable of capturing semantic similarity. In addition, we evaluated \textbf{cosine similarity}, a commonly used metric for embedding-based retrieval that measures angular similarity between vectors and is invariant to vector magnitude.

For document ranking, we evaluated an adaptive retrieval reranking, \textbf{CrossEncoder-based reranking} approach \cite{rozonoyer2026autoregressiverankingbridginggap}. Unlike bi-encoder retrieval, which embeds queries and documents independently, a CrossEncoder jointly encodes the query--document pair within a single transformer model and directly predicts a relevance score. While this approach yields more accurate relevance estimates, it incurs substantially higher computational overhead and is therefore applied only as a reranking step.

To study the impact of retrieval corpus composition, we constructed two retrieval corpora. The \textbf{documentation corpus} contained the Qiskit~2.0.1 API reference \cite{qiskit_api_201} and Qiskit~2.0 release notes \cite{qiskit_release_20}. The \textbf{code corpus} included the core Qiskit source code corresponding to version~2.2.0, extended with the \textit{qiskit-nature} \cite{qiskit_nature}, \textit{qiskit-machine-learning} \cite{qiskit_machine_learning}, \textit{qiskit-dynamics} \cite{qiskit_dynamics}, and \textit{qiskit-experiments} \cite{qiskit_experiments} repositories. We compared retrieval using documentation alone with retrieval over the combined documentation and source code corpora.

Retrieved content was added to the task prompt as additional contextual information before generation.

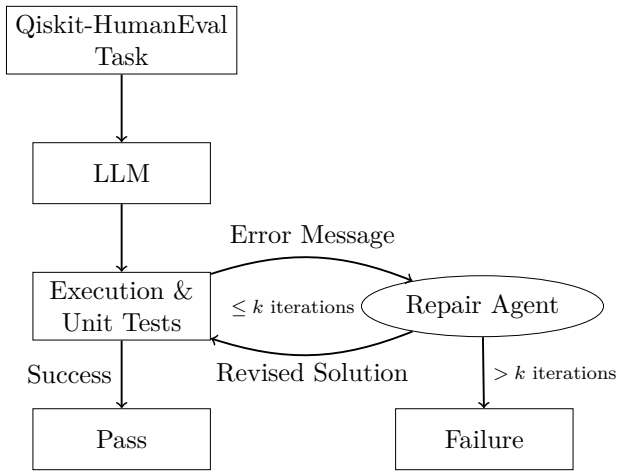
\begin{figure}[t]
\centering
\resizebox{\columnwidth}{!}{%
\begin{tikzpicture}[
  box/.style={draw, rectangle, align=center, minimum width=2.6cm, minimum height=0.9cm},
  oval/.style={draw, ellipse, align=center, minimum width=2.8cm, minimum height=0.9cm},
  arrow/.style={->, thick},
]
\node[box] (task) {Qiskit-HumanEval\\Task};
\node[box, below=of task] (llm) {LLM};
\node[box, below=of llm] (exec) {Execution \&\\Unit Tests};

\node[oval, right=2.2cm of exec] (agent) {Repair Agent};

\node[box, below=of exec] (pass) {Pass};
\node[box, right=2.7cm of pass] (fail) {Failure};

\draw[arrow] (task) -- (llm);
\draw[arrow] (llm) -- (exec);

\draw[arrow] (exec) -- node[left]{Success} (pass);
\draw[arrow] (exec) to[bend left=20] node[above]{Error Message} (agent);

\draw[arrow] (agent) -- node[right]{\scriptsize $> k$ iterations}(fail);
\draw[arrow] (agent) to[bend left=20] node[above, yshift=-15pt]{Revised Solution} (exec);

\node[left=1cm of agent, above, , yshift=-8pt] {\scriptsize $\leq k$ iterations};

\end{tikzpicture}%
}
\caption{Agent-based inference with execution feedback on the Qiskit-HumanEval benchmark.}
\label{fig:agentic_loop}
\end{figure}

\paragraph{Agent-based inference with execution feedback.}
Figure~\ref{fig:agentic_loop} illustrates the agent-based inference procedure, in which execution feedback from the Qiskit-HumanEval test harness is incorporated into the generation process. In this configuration, the model first generates a candidate solution, which is executed against the unit tests. If execution fails, the resulting Python error message or failed assertion is extracted and provided back to the model as feedback, prompting the generation of a revised solution. Only the error message is passed back to the model; the original prompt and any retrieved context remain unchanged, and no ground-truth information is revealed. To control computational cost, we evaluate bounded configurations with one to five repair attempts. If all attempts fail, the task is marked as unsuccessful.

\paragraph{Retrieval and scoring strategies.}
To study the impact of retrieval design on Qiskit code generation, we evaluated multiple retrieval and scoring strategies within the retrieval-augmented generation (RAG) framework.

We considered both single-corpus and combined-corpus retrieval settings. In the single-corpus setting, retrieval was performed exclusively over the documentation corpus. In the combined-corpus setting, retrieval was performed jointly over the documentation and Qiskit source code corpora.

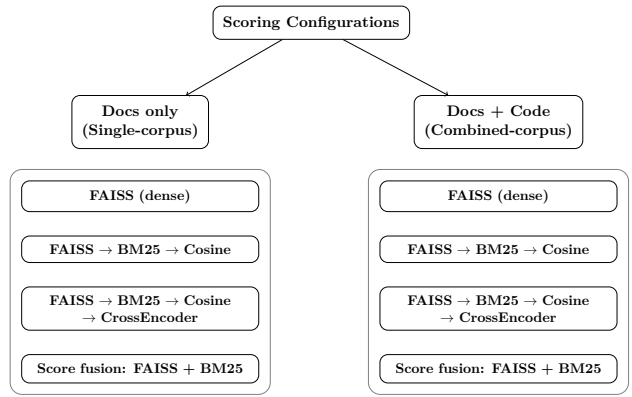
\begin{figure}[t]
\centering
\resizebox{\linewidth}{!}{%
\begin{tikzpicture}[
  font=\small,
  head/.style={draw, rounded corners=2mm, align=center, inner sep=7pt, font=\bfseries},
  box/.style={draw, rounded corners=2mm, align=center, inner sep=6pt, text width=55mm},
  arrow/.style={->, line width=0.45pt},
  group/.style={draw, rounded corners=3mm, line width=0.6pt, gray, inner sep=8pt}
]

\node[head] (root) {Scoring Configurations};

\node[head, below left=14mm and 1mm of root] (docs)
  {Docs only\\(Single-corpus)};

\node[head, below right=14mm and 1mm of root] (dual)
  {Docs + Code\\(Combined-corpus)};

\draw[arrow] (root) -- (docs);
\draw[arrow] (root) -- (dual);

\node[box, below=8mm of docs] (d1) {\textbf{FAISS (dense)}};
\node[box, below=6mm of d1] (d2) {\textbf{FAISS $\rightarrow$ BM25 $\rightarrow$ Cosine}};
\node[box, below=6mm of d2] (d3) {\textbf{FAISS $\rightarrow$ BM25 $\rightarrow$ Cosine}\\\textbf{$\rightarrow$ CrossEncoder}};
\node[box, below=6mm of d3] (d4) {\textbf{Score fusion: FAISS + BM25}};

\node[group, fit=(d1)(d2)(d3)(d4)] {};

\node[box, below=8mm of dual] (c1) {\textbf{FAISS (dense)}};
\node[box, below=6mm of c1] (c2) {\textbf{FAISS $\rightarrow$ BM25 $\rightarrow$ Cosine}};
\node[box, below=6mm of c2] (c3) {\textbf{FAISS $\rightarrow$ BM25 $\rightarrow$ Cosine}\\\textbf{$\rightarrow$ CrossEncoder}};
\node[box, below=6mm of c3] (c4) {\textbf{Score fusion: FAISS + BM25}};

\node[group, fit=(c1)(c2)(c3)(c4)] {};

\end{tikzpicture}%
}
\caption{RAG Retrieval: Configuration space of retrieval/scoring strategies evaluated, grouped by corpus indexing scheme.}
\label{fig:rag_config_space}
\end{figure}

For retrieval and ranking, we evaluated several approaches (see Figure~\ref{fig:rag_config_space}):
(i) dense retrieval using FAISS over sentence-level embeddings, 
(ii) sparse retrieval using BM25 keyword matching, 
(iii) cosine similarity–based scoring, 
(iv) cross-encoder re-ranking using a BGE-based reranker, and 
(v) hybrid score fusion strategies combining dense and sparse signals.

For dense retrieval, we varied the number of retrieved chunks and similarity metrics, and we explored different weighting schemes when combining dense and sparse scores.

Based on preliminary ablation experiments, we identified the best-performing retrieval configuration and adopted it for the main evaluation., as discussed in Section~\ref{sec:results}.

\section{Results}
\label{sec:results}

This section reports the results obtained across all model configurations evaluated in this study. As described in Section~\ref{sec:methods}, we tested three inference settings:
(i) \emph{zero-shot generation}, where the model receives only the task prompt;
(ii) \emph{retrieval-augmented generation (RAG)}~\cite{lewis2021retrievalaugmentedgenerationknowledgeintensivenlp}, where relevant Qiskit documentation and/or code is retrieved and added to the prompt; and
(iii) \emph{agent-based inference}, which extends RAG by iteratively refining model outputs using execution feedback.

Table~\ref{tab:overall_results} summarizes Pass@1 accuracy and total execution time across model families and inference strategies.


\begin{table*}[t!]
\centering
\small
\setlength{\tabcolsep}{5pt}
\renewcommand{\arraystretch}{1.15}
\begin{tabular}{lcccccccc}
\toprule
\multirow{2}{*}{Model} &
\multicolumn{4}{c}{Pass@1 (\%)} &
\multicolumn{4}{c}{Execution time (s)} \\
\cmidrule(lr){2-5}\cmidrule(lr){6-9}
& Zero-shot & RAG & Agent (1) & Agent (5) & Zero-shot & RAG & Agent (1) & Agent (5) \\
\midrule
\multicolumn{9}{l}{\textbf{Parameter-specialized baseline}} \\
Param-Spec.\ (Dupuis et al.) & 46.5 & -- & -- & -- & -- & -- & -- & -- \\
\midrule

\multicolumn{9}{l}{\textbf{OpenAI}} \\
GPT 5.2-codex       & 60.3 & 64.2 & 70.2 & 80.1 & 1{,}402 & 1{,}424 & 2{,}751 & 5{,}713 \\
GPT 5.2             & 61.6 & 60.3 & 73.5 & 81.4 & 1{,}486 & 1{,}514 & 2{,}637 & 5{,}916 \\
GPT 5               & 60.3 & 64.2 & 72.2 & 80.8 & 2{,}285 & 2{,}285 & 3{,}188 & 9{,}037 \\
GPT o3                  & 53.6 & 57.0 & 69.5 & 79.5 & 1{,}622 & 1{,}622 & 2{,}344 & 6{,}123 \\
GPT 4o              & 44.4 & 46.4 & 58.9 & 65.6 &   423  &   423  &   639  & 1{,}395 \\
GPT 4.1             & 47.0 & 51.7 & 64.9 & 74.2 &   450  &   450  & 1{,}135 & 1{,}424 \\
GPT-4.1 + GPT-5     &  --  &  --  & 70.2 & 79.5 &   --   &   --   & 2{,}073 & 8{,}091 \\
\midrule

\multicolumn{9}{l}{\textbf{Claude}} \\
Opus 4.6     & 68.9 & 70.9 & 80.8 & 85.4 &   582  &   725  &   974  & 1{,}363 \\
Sonnet 4.6    & 69.5 & 64.9 & 79.5 & 81.5 &   1267 &   744  &   1{,}564  & 2{,}176 \\
Sonnet 4.5    & 61.6 & 57.6 & 62.9 & 76.2 &   829  &   896  & 1{,}182 & 1{,}683 \\
Haiku 4.5    & 49.7 & 51.7 & 64.2 & 70.9 &   296  &   385  &   509  & 1{,}127 \\
Haiku 3.5     & 42.4 & 41.7 & 45.0 & 56.9 &   408  &   527  &   712  & 1{,}533 \\
\midrule

\multicolumn{9}{l}{\textbf{Gemini}} \\
Pro 3                & 68.9 & 64.2 & 74.2 & 75.5 & 3{,}835 & 4{,}737 & 7{,}561 & 12{,}312 \\
Pro 2.5              & 54.3 & 54.3 & 60.6 & 65.6 & 3{,}149 & 3{,}149 & 10{,}904 & 10{,}963 \\
Flash 3              & 61.6 & 61.6 & 70.9 & 76.8 & 2{,}847 & 3{,}202 & 4{,}867  & 9{,}658 \\
Flash 2.5            & 45.1 & 49.1 & 55.7 & 66.2 & 1{,}294 & 1{,}294 & 3{,}351  & 5{,}246 \\
\bottomrule
\end{tabular}

\caption{\textbf{Overall Pass@1 accuracy and execution time} across model families and inference strategies on \texttt{Qiskit HumanEval} (151 tasks). Execution time is the total time over the full evaluation set. Agent (1) denotes a single generate--execute--repair step; Agent (5) allows up to five repair attempts. The parameter-specialized baseline reports Pass@1 only (runtime not available).}
\label{tab:overall_results}
\end{table*}


We first describe the selection of RAG retrieval and scoring strategies, which required the most extensive configuration analysis. We then present benchmark results comparing the tested model families.

\subsection{RAG Retrieval and Scoring Strategy Selection}
\label{sec:rag_scoring_results}

Each inference setting involves multiple design choices. Among them, Retrieval-Augmented Generation (RAG) required the most careful tuning to identify a stable and effective configuration. Figure~\ref{fig:rag_pipeline} illustrates the RAG pipeline variants evaluated in our experiments.

All retrieval ablation experiments in this section were conducted on a subset of OpenAI models.

\paragraph{Single-corpus vs. combined-corpus retrieval.}
We first evaluated retrieval using the documentation corpus alone, based solely on Qiskit documentation. Across all tested scoring strategies, this setup did not yield consistent improvements in Pass@1 accuracy. 

In contrast, retrieval over the combined \emph{documentation} and \emph{source code} corpora produced measurable gains. Using dense embedding-based retrieval with FAISS and a small retrieval depth, we observed improvements of approximately $1$--$4$ absolute percentage points in Pass@1 relative to the zero-shot baseline without retrieval.

\paragraph{Retrieval depth.}
The most effective configuration used dense FAISS retrieval with a low retrieval depth ($k = 4$). Increasing $k$ to larger values (e.g., $k \in \{10, 20, 30\}$) consistently degraded performance, suggesting that additional retrieved context introduced noise and reduced the relevance of the augmented prompt.

\paragraph{Scoring strategies and reranking.}
In preliminary tuning experiments on a representative subset of OpenAI models, more complex retrieval pipelines, including FAISS $\rightarrow$ BM25 $\rightarrow$ cosine similarity,
or FAISS $\rightarrow$ BM25 $\rightarrow$ cosine similarity $\rightarrow$ CrossEncoder reranking (BGE-reranker-large), did not yield consistent improvements over dense FAISS alone. CrossEncoder reranking additionally introduced noticeable retrieval latency during evaluation, reducing its practical appeal for large-scale experiments.

\paragraph{Score fusion.}
Hybrid score fusion between FAISS and BM25 produced limited and inconsistent gains in our tuning experiments. Equal-weight fusion did not lead to observable improvements, while assigning a higher weight to FAISS (approximately $2\times$) occasionally yielded modest gains. Given the small and unstable improvements relative to the added complexity, score fusion was not adopted in the default configuration.

\paragraph{Embedding models.}
Across preliminary ablation experiments, \textsl{text-embedding-3-large} consistently yielded the strongest retrieval performance in terms of downstream pass@1 accuracy. Although smaller embedding models offered lower computational cost, the larger model provided more reliable semantic matching between task prompts and relevant Qiskit documentation. Consequently, \textsl{text-embedding-3-large} was adopted for the main evaluation.

\subsubsection{Effect of Retrieval Corpus Composition}
\label{sec:index_results}

We further analyzed the impact of retrieval corpus composition on RAG performance. For the code corpus, we compared a \emph{core-only} corpus built only from the Qiskit v2.2.0 repository against a larger multi-repository corpus including \texttt{qiskit-nature}, \texttt{qiskit-machine-learning}, \texttt{qiskit-dynamics}, and \texttt{qiskit-experiments}.

Across tested OpenAI configurations, the core-only corpus consistently outperformed the larger multi-repository corpus. While the extended corpus increased overall coverage, it also introduced less relevant code fragments for the benchmark tasks, reducing the average relevance of retrieved context and negatively affecting generation quality.

As a result, the final RAG configuration uses a combined-corpus setup consisting of Qiskit documentation and core Qiskit source code only.

\subsubsection{Key Findings for RAG}
\label{sec:rag_key_findings}

Our ablation study leads to the following conclusions:

\begin{itemize}
		\item Among the evaluated embedding models, \textsl{text-embedding-3-large} consistently yielded the strongest downstream pass@1 performance.
    \item Dense retrieval using FAISS with a small retrieval depth is the most effective RAG strategy in our setting.
    \item Combining documentation and code corpora yields better results than documentation-only retrieval.
    \item Sparse BM25 retrieval provides little additional benefit for code-centric tasks.
    \item CrossEncoder reranking is computationally expensive and does not improve generation quality in this benchmark.
\end{itemize}


\begin{figure*}[t]
    \centering
    \includegraphics[width=1.05\textwidth]{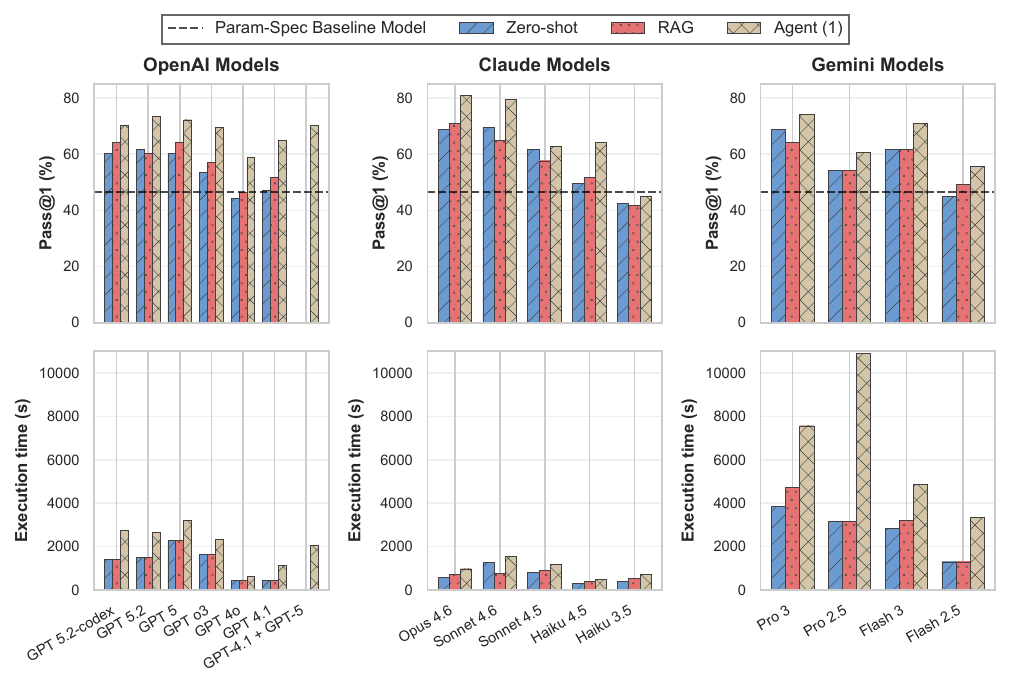}
    \caption{\textbf{General accuracy (top) and execution time (bottom) under different specialization strategies.}
We compare a training-time \emph{parameter-specialized} baseline model (Param-Spec.; fine-tuned Granite reported by Dupuis et al.~\cite{dupuis2024qiskitcodeassistanttraining}) against general-purpose LLMs evaluated with inference-time system-level specialization: zero-shot, retrieval-augmented generation (RAG), and a single-step generate--execute--repair loop (Agent).}
\label{fig:grouped_asymmetric}
\end{figure*}

\begin{figure*}[t]
    \centering
    \includegraphics[width=1.05\textwidth]{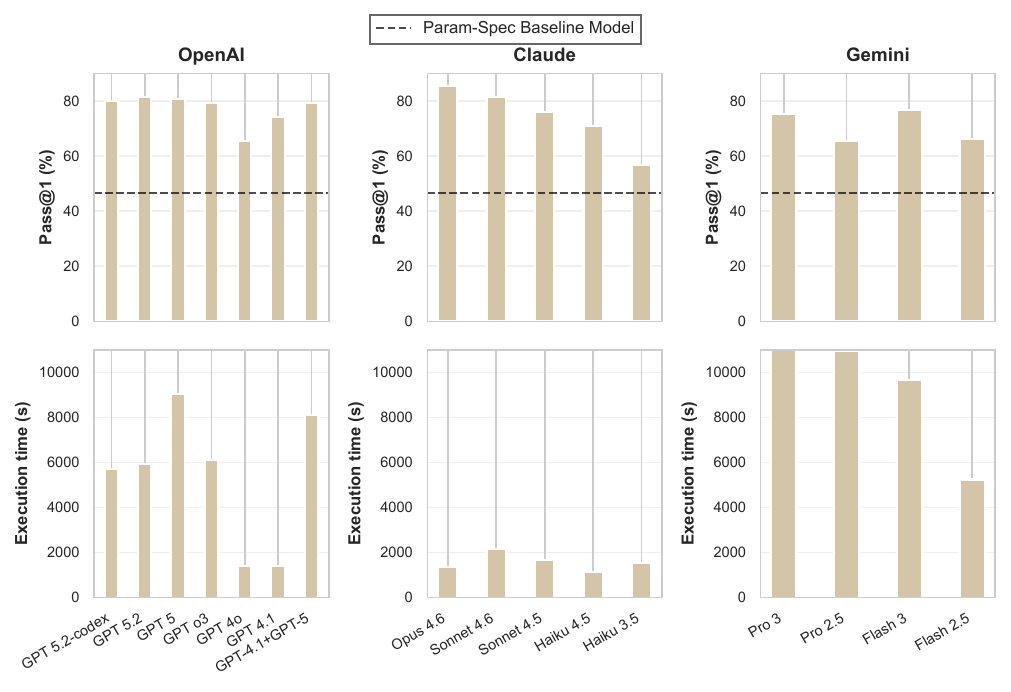}
    \caption{\textbf{General accuracy (top) and execution time (bottom) across model families under multi-step agentic inference with up to five repair attempts.}
Results correspond to iterative generate--execute--repair loops, where models attempt to correct failed executions using unit-test feedback until success or the maximum number of repair attempts is reached. Execution time is measured over the full evaluation set of 151 benchmark tasks.}
\label{fig:five_repair_attempts}
\end{figure*}

\begin{figure*}[t]
    \centering
    \includegraphics[width=1.05\textwidth]{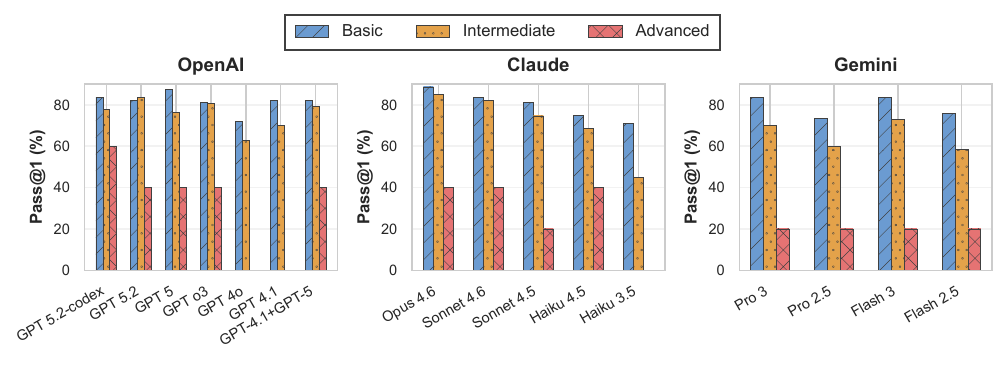}
    \caption{\textbf{Pass@1 accuracy by problem difficulty tier (basic, intermediate, advanced) across model families under multi-step agentic inference with up to five repair attempts.}
Grouped bars within each model family report tier-specific accuracy on the \texttt{Qiskit HumanEval} benchmark, enabling direct comparison of performance as task difficulty increases.}
\label{fig:accuracy_by_tier_families}
\end{figure*}


\subsection{Model Benchmarking}
\label{sec:model_benchmarks}

We evaluate a diverse set of large language models from OpenAI \cite{openai_models_2025}, Anthropic (Claude) \cite{anthropic_claude_models_2025}, and Google (Gemini) \cite{geminiteam2025geminifamilyhighlycapable} on the \texttt{Qiskit HumanEval} benchmark \cite{qiskit_humaneval_2024}, and compare them against a training-time parameter-specialized baseline (the fine-tuned Granite model reported by Dupuis et al.~\cite{dupuis2024qiskitcodeassistanttraining}). All results are reported for the best-performing configuration of each inference setting. We analyze both functional correctness (measured as Pass@1) and execution time, highlighting accuracy--runtime trade-offs across inference strategies.

\paragraph{Baseline, RAG, and single-step agentic inference.}
Figure~\ref{fig:grouped_asymmetric} (top row) reports general accuracy under three inference settings: zero-shot prompting, RAG, and agentic inference with a single execution--repair loop. The bottom row shows the corresponding execution times. Execution time is not reported for the Granite baseline, as runtime measurements were not provided in the original evaluation study \cite{dupuis2024qiskitcodeassistanttraining}.

We observe that modern foundation models already match or exceed the performance of the parameter-specialized baseline, despite being evaluated without task-specific fine-tuning, RAG, or agentic inference. For OpenAI reasoning-capable models, we evaluated \texttt{reasoning\_effort} in \{\texttt{minimal}, \texttt{low}, \texttt{medium}, \texttt{high}\} and compared verbosity settings. Across configurations, \texttt{reasoning\_effort=low} combined with \texttt{verbosity=low} (default verbosity is \texttt{medium}) generally provides a more favorable accuracy--runtime trade-off, substantially reducing execution time while maintaining comparable or improved accuracy for most models. Table~\ref{tab:openai_param_sensitivity} reports representative results comparing the default (\texttt{medium/medium}) and selected (\texttt{low/low}) configurations. These findings further support the competitiveness of general-purpose models without task-specific fine-tuning.

Across model families, RAG does not consistently improve accuracy. While it provides modest gains in some cases, several models exhibit neutral or degraded performance relative to zero-shot inference. In contrast, agentic inference with execution feedback improves accuracy for all evaluated models. In this setting, the model first generates a candidate solution, executes it against the unit tests, and—if execution fails—receives the resulting Python error message or failed assertion as feedback for a revised generation.

This accuracy improvement comes at a significant runtime cost. Execution time increases substantially for agentic inference, particularly for \texttt{Gemini Pro~3} and \texttt{Gemini Pro~2.5}, where runtime overhead grows sharply. By comparison, RAG maintains execution times close to the zero-shot baseline.

All agentic runs in Figure~\ref{fig:grouped_asymmetric} use the same model for both generation and repair, except for the \texttt{GPT-4.1 + GPT-5} configuration. In this hybrid setup, \texttt{GPT-4.1} generates the initial solution, while \texttt{GPT-5} performs the repair step when execution fails. This approach improves performance relative to using \texttt{GPT-4.1} alone, though it remains slightly worse in accuracy than running the full agentic loop entirely with \texttt{GPT-5}.

For OpenAI models, retrieval-augmented generation (RAG) generally improves pass@1 performance relative to zero-shot generation, with gains observed across most evaluated versions. In contrast, Claude and Gemini models exhibit mixed behavior under RAG, with several variants showing marginal improvements and others experiencing slight performance degradation. This suggests that, under the default configuration used in our experiments, OpenAI models may integrate retrieved documentation more effectively than Claude and Gemini models.

\paragraph{Multi-step agentic inference.}
Figure~\ref{fig:five_repair_attempts} isolates agentic inference with up to five repair attempts. In this configuration, the agent iteratively generates, executes, and repairs code until either all tests pass or the maximum number of attempts is reached.

Across all evaluated model families, multi-step agentic inference consistently matches or exceeds the parameter-specialized baseline, with several models exceeding 75\% pass@1 accuracy. These results demonstrate that inference-time system-level augmentation can surpass parameter-level fine-tuning on this benchmark.

However, these gains come with additional computational cost. Runtime generally increases with additional repair attempts, particularly for Gemini and some OpenAI models. In contrast, Claude models exhibit comparatively efficient scaling; notably, Opus 4.6 under deeper agentic loops remains highly competitive in execution time, and in some cases even runs faster than zero-shot baselines of certain OpenAI and Gemini variants. Overall, this suggests that the accuracy–runtime trade-off under increasing agent depth is highly model-dependent rather than uniformly prohibitive.

To prevent rare long-running or non-terminating executions from dominating runtime measurements, we enforced a per-attempt timeout of 10 minutes when running the Qiskit-HumanEval test harness. Timeouts can occur on tasks that involve randomized search or unbounded loops, and were observed primarily for Gemini Pro 3 in our experiments. If a timeout was reached, the attempt was terminated, marked as failed and the agent proceeded to the next repair attempt (or stopped if the maximum number of attempts was reached).

\paragraph{Accuracy by difficulty tier.}

The \texttt{Qiskit HumanEval} \cite{qiskit_humaneval_2024} benchmark assigns each problem to one of three difficulty levels: \emph{basic}, \emph{intermediate}, and \emph{advanced}. Figure~\ref{fig:five_repair_attempts} reports aggregate performance across all difficulty tiers, without separating results by problem level. Figure~\ref{fig:accuracy_by_tier_families} decomposes the multi-step agentic results (agentic inference with up to five repair attempts) according to these levels, thereby highlighting how model performance varies as problem difficulty increases.

As expected, accuracy decreases sharply with increasing difficulty. OpenAI models demonstrate strong performance on advanced problems, with several models solving approximately 40\% of advanced tasks (2 out of 5). Claude models show comparable behavior, with top variants such as \texttt{Opus 4.6} and \texttt{Sonnet 4.6} achieving similar performance levels on advanced tasks. In contrast, Gemini models exhibit lower performance on advanced problems, consistently solving around 20\% (1 out of 5) of advanced tasks.

Overall, these tiered results indicate that while multi-step agentic inference substantially improves performance across all families, advanced tasks remain significantly more challenging across all model families.


\section{Analysis and Discussion}

The benchmarking results demonstrate that modern general-purpose LLMs can achieve strong performance on Qiskit code generation tasks even without domain-specific fine-tuning. This represents a notable shift from earlier approaches that relied heavily on parameter-level specialization to encode framework-specific knowledge.

A key contributing factor is the improved reasoning, structured code generation, and error recovery capabilities of recent modern general-purpose models. As shown in Table~\ref{tab:overall_results}, these models are increasingly able to generate syntactically correct and semantically meaningful quantum code that satisfy nontrivial unit-test constraints. This progress reduces the relative advantage of static fine-tuning, particularly for tasks that require structured reasoning over API interactions rather than memorization of domain-specific patterns.

Inference-time adaptation further amplifies these capabilities. Retrieval-augmented generation (RAG) and execution-based feedback externalize domain knowledge that would otherwise need to be embedded in model parameters. Instead of encoding Qiskit APIs and documentation through fine-tuning, models can dynamically access relevant information and correct errors at generation time. While RAG does not consistently improve accuracy across all model families, it generally preserves runtime efficiency and can provide targeted benefits depending on the task and model. Agentic inference, by contrast, reframes code generation as an iterative reasoning process and yields substantial accuracy gains for models capable of effectively leveraging execution feedback.

\subsection{Specialization Strategies: Fine-Tuning versus Inference-Time Adaptation}

These results highlight a broader shift in the trade-offs between parameter-level fine-tuning and inference-time adaptation. Fine-tuning remains an effective mechanism for specialization, but it is inherently static: once trained, a model’s knowledge is fixed and must be updated through additional training cycles as APIs and software ecosystems evolve.

Inference-time adaptation mechanisms—such as retrieval and agentic execution—offer greater flexibility. They allow models to incorporate updated documentation, new APIs, and evolving execution semantics without retraining. In the context of Qiskit code generation, our results show that general-purpose models equipped with inference-time augmentation can match or exceed the performance of a fine-tuned baseline, while significantly reducing the need for repeated retraining as the framework evolves.

This distinction is particularly important in fast-moving domains such as quantum software, where SDK changes are frequent and backward compatibility is limited. In such settings, inference-time adaptation provides a more sustainable path to maintaining high performance over time.

\subsection{Deployment and Maintenance Implications}

Beyond accuracy, inference-time adaptation has important practical implications for deployment and maintenance. Fine-tuning typically requires training and maintaining a separate specialized model for each base architecture, which introduces significant overhead when models are updated or replaced. In contrast, retrieval and agentic workflows can often be applied using a shared configuration across multiple underlying models.

This decoupling of model capability from domain specialization reduces engineering effort and deployment complexity. A single agentic or retrieval pipeline can be reused across different model families, enabling rapid experimentation, easier model swapping, and faster adoption of new frontier models. From a production perspective, this substantially lowers the cost and time required to maintain effective quantum programming assistants, while improving robustness to model and SDK updates.

\subsection{Inference-Time Trade-Offs}

Inference-time adaptation is not without cost. Agentic configurations require multiple generation–execution cycles, increasing the number of model calls and overall inference-time compute. We therefore report execution time alongside accuracy to contextualize these trade-offs. The results indicate that accuracy improvements from agentic inference come at a nontrivial increase in runtime, particularly for larger models.

While some latency overhead can be mitigated through parallelization or optimized implementation, agentic methods inherently involve deeper inference pipelines than single-pass or retrieval-augmented approaches. By contrast, retrieval-augmented generation maintains execution times close to the zero-shot baseline while still providing measurable accuracy gains.

Comparable inference-time measurements are not available for the parameter-specialized baseline, as runtime data were not reported in the original evaluation study~\cite{dupuis2024qiskitcodeassistanttraining}. Nevertheless, the observed trends illustrate a practical design consideration: higher accuracy through multi-step inference is achieved at the expense of increased runtime, and selecting an inference strategy therefore depends on application-specific performance and efficiency requirements.


\section{Limitations of the Benchmarking Study}

Despite the encouraging results presented in this work, several limitations related to benchmarking methodology, evaluation validity, and reproducibility should be acknowledged.

\subsection{Benchmarking and Evaluation Considerations}

Benchmarking code generation models in specialized domains involves methodological challenges that extend beyond raw accuracy metrics. Differences in inference configurations, tool availability, execution environments, and agent design choices can substantially influence observed performance. Consequently, direct comparisons between fine-tuned models and inference-time–augmented systems may conflate modeling capability with evaluation setup.

These considerations highlight the importance of transparent reporting of inference settings, tooling assumptions, and evaluation protocols. Without such clarity, performance differences attributed to model architecture or training may instead reflect differences in inference-time instrumentation.

\subsection{Benchmark Exposure and Leakage Risk}

The \texttt{Qiskit HumanEval} benchmark \cite{qiskit_humaneval_2024} is publicly available and closely aligned with Qiskit documentation, tutorials, and open-source repositories. As a result, it is not possible to rule out prior exposure to benchmark tasks --or closely related implementations-- during large-scale pretraining of frontier models trained on web-scale corpora.

Consequently, while prior exposure to benchmark tasks or related code cannot be entirely excluded, there is no direct evidence of task-level memorization in our evaluation. Given the structured nature of the benchmark and the requirement for test-passing solutions, strong performance likely reflects a combination of learned coding patterns and task-level reasoning. This consideration applies broadly to publicly available code benchmarks and does not necessarily indicate benchmark-specific contamination.

The risk is further amplified in retrieval-augmented and agent-based configurations, where retrieved documentation or execution feedback may implicitly constrain the solution space. Although filtering is applied to mitigate direct solution leakage, fully eliminating benchmark exposure remains an open challenge in the evaluation of LLM-based code generation systems.

\subsection{Non-Determinism and Evaluation Variance}

LLM inference is inherently stochastic, and this variability is amplified in agent-based workflows involving multiple generation and execution steps. While Pass@k-style metrics mitigate some of this variance, results may still vary across runs depending on decoding settings and stochastic sampling behavior.

\subsection{Model Versioning and Availability}

Although commercial LLM APIs expose explicit model version identifiers, long-term reproducibility remains constrained by model lifecycle policies, updates, and potential deprecation. Consequently, the reported results should be interpreted as time-indexed performance snapshots rather than permanent benchmarks. While open-weight models would be preferable for archival reproducibility, they currently lag behind frontier proprietary models in performance on this task.

\subsection{Cost and Computational Overhead}

From a benchmarking perspective, the evaluated inference strategies are not cost-normalized. Agent-based configurations, particularly those involving multiple execution--repair cycles, operate under substantially larger inference-time compute budgets than single-pass or retrieval-augmented settings.

Consequently, the reported accuracy improvements reflect performance under unequal computational conditions. Although execution time is reported to contextualize this overhead, we do not constrain models to fixed time, token, or compute budgets. A stricter cost-aware evaluation—such as comparing methods under equalized computational budgets—would provide a more controlled assessment of efficiency and remains an important direction for future benchmarking work.

\subsection{Scope of Generalization}

Finally, our evaluation is limited to Qiskit and the \texttt{Qiskit HumanEval} benchmark \cite{qiskit_humaneval_2024}. While the observed trends are likely indicative of broader patterns in scientific and API-driven programming tasks, further studies are required to assess generalization across different quantum SDKs, programming paradigms, and benchmark designs.


\section{Conclusion}

In this work, we conducted a comprehensive benchmarking study of large language models for Qiskit code generation, comparing modern general-purpose LLMs against a domain-specific fine-tuned baseline. Our evaluation spans multiple inference strategies—including zero-shot prompting, retrieval-augmented generation, and agentic inference with execution feedback—and analyzes both functional correctness and execution-time behavior on the \texttt{Qiskit HumanEval} benchmark.

Our results demonstrate that recent general-purpose LLMs can match or surpass the performance of a fine-tuned baseline without requiring domain-specific training, thereby eliminating the need for additional training compute and reducing the maintenance burden associated with repeated re-specialization as frameworks evolve.
Inference-time adaptation plays a central role in this shift: while retrieval-augmented generation (RAG) yields modest and model-dependent gains, agentic inference with execution feedback leads to substantial accuracy improvements for models capable of iterative self-correction. In particular, OpenAI models consistently benefit from execution feedback across multiple repair iterations. Claude models also exhibit accuracy gains from deeper agentic loops while maintaining relatively stable execution times. Gemini models similarly improve under agentic refinement; however, their runtime scaling varies more substantially across variants, with some configurations incurring noticeably higher overhead for comparable accuracy gains.

These accuracy gains come at the cost of increased inference-time compute, as repeated generation–execution cycles require additional model calls and test executions. From a benchmarking perspective, this introduces an accuracy–efficiency trade-off: agentic methods consume more inference-time resources per task than retrieval-augmented or single-pass approaches.

However, in absolute terms, the additional runtime and API costs observed in our experiments remain moderate and may be acceptable in many practical deployment scenarios, particularly when accuracy improvements are mission-critical. While some latency overhead can be mitigated through parallelization and optimized orchestration, agentic configurations inherently involve deeper inference pipelines. Developing cost-aware and budget-normalized evaluation protocols—such as comparing methods under equalized time or token budgets—remains an important direction for future benchmarking work.

Among all evaluated configurations, the model Opus 4.6 under five-step agentic inference achieved the highest observed accuracy (85.4\% pass@1), substantially surpassing the parameter-specialized baseline (46.5\%) and representing the strongest performance measured in this study. Notably, this configuration maintained competitive total execution time (1,363 seconds over the full benchmark, corresponding to approximately 9 seconds per task), indicating that substantial accuracy gains from execution-feedback loops do not necessarily entail prohibitive runtime overhead.

Beyond raw performance, our study underscores broader implications for the design of quantum software tooling. Inference-time adaptation enables a decoupling between model capability and domain specialization, allowing a single agentic or retrieval pipeline to be reused across multiple underlying models. This reduces the engineering and maintenance burden associated with repeated fine-tuning and offers a more flexible and sustainable approach for rapidly evolving quantum SDKs such as Qiskit.

Overall, our findings suggest that inference-time augmentation—rather than parameter-level specialization—provides a practical and adaptable pathway for building LLM-based quantum programming assistants. More broadly, this reflects a shift in LLM-assisted software development toward tool-augmented and execution-aware systems that better accommodate rapidly evolving scientific ecosystems. Future work should explore cost- and budget-normalized benchmarking, extend evaluations to additional quantum software frameworks, and investigate agent designs that more effectively balance reasoning depth with computational efficiency.


\section*{Acknowledgements}

The authors would like to thank the authors of the Qiskit Code Assistant study \cite{dupuis2024qiskitcodeassistanttraining} for making the Qiskit-HumanEval benchmark \cite{qiskit_humaneval_2024} publicly available and for providing a clear and reproducible evaluation protocol. Open access to benchmarks and reference results is essential for transparent and cumulative progress in LLM-assisted quantum software development, and this work would not have been possible without those contributions.


\bibliographystyle{quantum}
\bibliography{bibliographystyle}

\clearpage
\onecolumn
\appendix


\begin{table*}[t]
\centering
\small
\setlength{\tabcolsep}{6pt}
\renewcommand{\arraystretch}{1.15}
\begin{tabular}{lcccc}
\toprule
\multirow{2}{*}{Model} 
& \multicolumn{2}{c}{\texttt{reasoning\_effort=medium}, \texttt{verbosity=medium}} 
& \multicolumn{2}{c}{\texttt{reasoning\_effort=low}, \texttt{verbosity=low}} \\
\cmidrule(lr){2-3} \cmidrule(lr){4-5}
& Pass@1 (\%) & Time (s) & Pass@1 (\%) & Time (s) \\
\midrule
\multicolumn{5}{l}{\textbf{OpenAI reasoning-capable models}} \\
GPT-5        & 45.0 & 3594 & 50.9 & 2155 \\
GPT-5-mini   & 46.4 & 3420 & 47.7 & 1748 \\
GPT-5-nano   & 17.9 & 1716 & 36.4 & 1102 \\
o1-mini & 42.4 & 935 & \multicolumn{2}{c}{Not supported} \\
o3           & 53.0 & 2398 & 52.3 & 1600 \\
o3-mini      & 41.7 & 1541 & 37.8 & 1178 \\
o4-mini      & 42.4 & 1930 & 46.4 & 1201 \\
\midrule
\multicolumn{5}{l}{\textbf{OpenAI general-purpose models}} \\
GPT-4o        & 47.0 & 427 & 47.0 & 427 \\
GPT-4o-mini   & 30.5 & 387 & 30.5 & 387 \\
GPT-4.1       & 49.9 & 467 & 49.9 & 467 \\
GPT-4.1-mini  & 46.4 & 359 & 46.4 & 359 \\
GPT-4.1-nano  & 33.1 & 295 & 33.1 & 295 \\
\bottomrule
\end{tabular}
\caption{\textbf{OpenAI configuration sensitivity (reasoning\_effort, verbosity).}
Comparison of \texttt{reasoning\_effort} and \texttt{verbosity} settings on Pass@1 and execution time.
For reasoning-capable models, \texttt{reasoning\_effort=low} with \texttt{verbosity=low} generally provides a better accuracy--runtime trade-off.
The \texttt{low/low} configuration was not supported for \texttt{o1-mini}, as some non-default parameter combinations are not uniformly available across models.
General-purpose models are unaffected by these parameters.}
\label{tab:openai_param_sensitivity}
\end{table*}

\end{document}